\documentclass[letterpaper, 10 pt, conference]{ieeeconf}  

\IEEEoverridecommandlockouts                              

\usepackage{graphics} 
\usepackage{epsfig} 

\title{\LARGE \bf
Reducing the Amount of Real World Data for Object Detector Training
with Synthetic Data
}

\author{Sven Burdorf$^{1}$, Karoline Plum$^{2}$, and Daniel Hasenklever$^{1}$
\thanks{$^{1}$Sven Burdorf and Daniel Hasenklever are
        with dSPACE GmbH, Rathenaustr. 26, 33102 Paderborn, Germany
        {\tt\small sburdorf@dspace.de, dhasenklever@dspace.de}}%
\thanks{$^{2}$Karoline Plum is with the Institute of Computer Sciences,
        University Osnabr\"uck, 49069 Osnabr\"uck, Germany. Work was done while
        being an intern at dSPACE GmbH
        {\tt\small kplum@uni-osnabrueck.de}}%
}

\begin{document}

\maketitle
\thispagestyle{empty}
\pagestyle{empty}

\begin{abstract}
A number of studies have investigated the training of neural networks
with synthetic data for applications in the real world.
The aim of this study is to quantify how much real world data can be
saved when using a mixed dataset of synthetic and real world data.
By modeling the relationship between the number of training examples
and detection performance by a simple power law, we find that
the need for real world data can be reduced by up to 70\% without
sacrificing detection performance.
The training of object detection networks is especially enhanced
by enriching the mixed dataset with classes underrepresented
in the real world dataset.
The results indicate that mixed datasets
with real world data ratios between 5\% and 20\% reduce the need for real
world data the most without reducing the detection performance.
\end{abstract}

\section{INTRODUCTION}

Reliable and robust object detection is one of the key techniques for
autonomous driving.
However, the training of neural networks for object detection requires
a huge amount of labeled data for which collecting and labeling
in the real world is a time-consuming and expensive task.
Especially the collection of edge or corner cases for the dataset is
challenging, because it is virtually impossible to capture
all edge cases imaginable or even a significant amount of it.
A systematic overview of edge cases can be found in \cite{Breitenstein2020}.
Another problem of real world automotive datasets is
that there is usually a strong 
imbalance in the occurrence of classes: \emph{car} is the dominant class while
vulnerable road users like pedestrians and bicycles occur less frequently.
An additional challenge is the need for anonymization of real world data to
comply with the General Data Protection Regulation (GDPR).
To be compliant with the General Data Protection Regulation (GDPR) real
world data has to be usually anonymized, which adds a 
post-processing step to the data pipeline.

Synthetic data can be easily constructed in ways, that it has
none of the drawbacks mentioned above:
Ground truth data can be generated with pixel level accuracy.
It is even possible to generate these data "on the fly" without the need to
store the data and thus storage costs can be traded for compute costs
\cite{Mason2019}.
Even rare and dangerous scenarios can be simulated.
The class imbalance in synthetic datasets can also be circumvented by adding
additional instances of underrepresented classes to the synthetic dataset.
In addition, synthetic data does not suffer from any legal issues related to
the GDPR.
However, the domain gap prevents object detectors trained only 
on synthetic data from reaching the same performance as their counterparts trained
on an equal number of real training examples \cite{Chen2020, Chen2021}.

To overcome this problem mainly two different 
training strategies are usually applied:
One kind of approaches uses pretraining and fine-tuning strategies to
pretrain a network on synthetic data and later fine-tune on real data.
Another group of approaches is mixed strategies, where the networks are trained
simultaneously on real and synthetic data in a mixed training dataset.
Usually for both types of strategies the training does not start from scratch but from 
network backbones trained on real data as e.g. ImageNet \cite{Deng2009}
or COCO \cite{Lin2014}.

More specifically, several concrete approaches to combinine synthetic 
and real data for training have been applied so far.
For example
Trembaly et al. proposed pretraining on synthetic training examples, which are
generated by a domain randomization approach \cite{Trembaly2018}. Objects with
random textures are placed in random backgrounds with additional random
objects as distractors.
Fine-tuning these pretrained networks on real data improved the performance
of the object detection network compared to only training on real data.
However, their study focused on the dominant car class in automotive datasets
and did not deal with vulnerable road users.

As an example for the second group of strategies,
Nowruzi et al. studied the effect of different ratios of real and synthetic
data in the training dataset as well as pretraining and fine-tuning vs. a
mixed strategy \cite{Nowruzi2019}. They used various synthetic datasets
(Synscapes \cite{Wrenninge2018}, GTA \cite{Richter2017},
Carla \cite{Dosovitskiy2017})
and real datasets 
(BDD \cite{Yu2018}, Cityscapes \cite{Cordts2016}, 
KITTI \cite{Geiger2012}, Nuscenes \cite{Caesar2020})
and concluded that pretraining on synthetic data and fine-tuning on real data
provides better results than mixed training.
Although their results showed, that higher ratios of real data in the training
dataset are beneficial, only ratios of real data up to 10\% were
investigated.

However, when choosing generally mixed training sets as basic strategy,
it is still unclear, what the best ratio of real and synthetic
data would be.
Also, the actual criteria, to get the "best" ratio can still be defined
in several ways.

This study focuses on determining how much real data can be saved when
using synthetic data.
Therefore, an evaluation method based on a
simple power law is proposed to quantify
real data saving in mixed real and synthetic datasets.
In addition, the influence of adding synthetic instances of
underrepresented vulnerable road users (\emph{person} class) is studied.

Apart from 3D rendering approaches to produce synthetic data,
generative adversarial networks (GANs) are a promising alternative
to generate synthetic data \cite{Frid-Adar2018, Park2019}.
To investigate the possibility of using GAN images for training neural
networks, an additional synthetic dataset is created from the Synscapes
semantic segmentation by an image-to-image translation GAN \cite{Wang2018}.
This dataset is referred to as "GANscapes".

\section{METHOD}

\begin{figure*}[t]
    \centering
    \parbox{5.5in}{
      \includegraphics[width=5.5in]{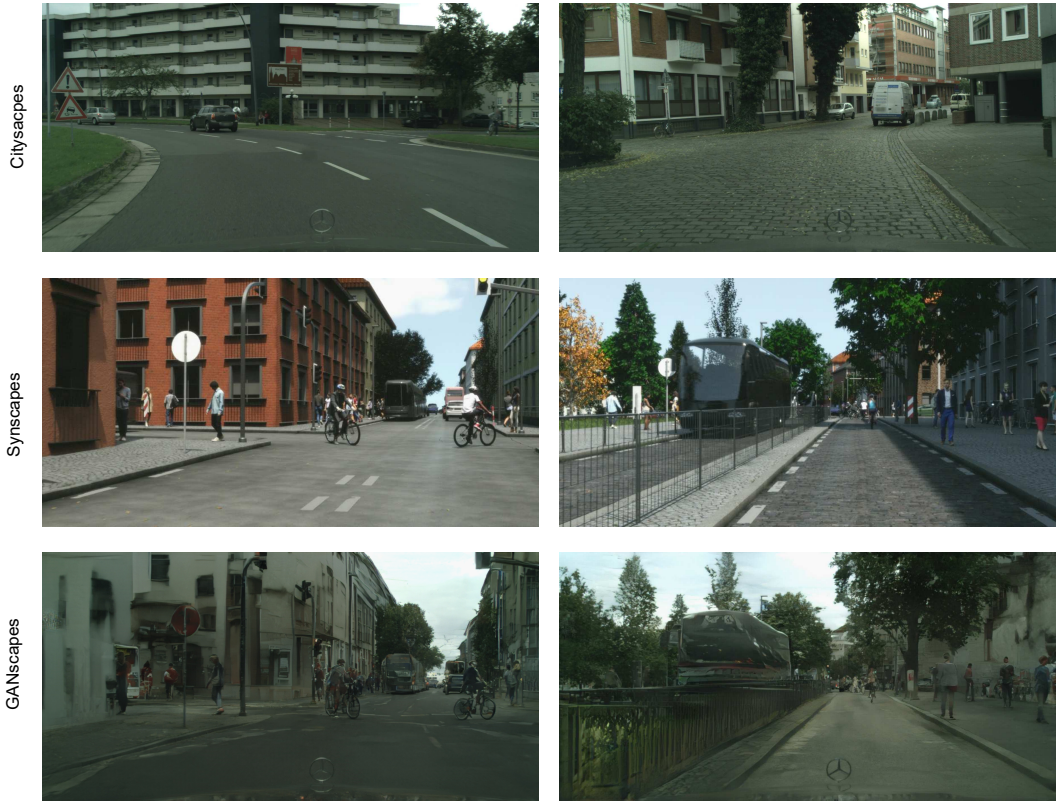}
    }
    \caption{Example images from the Cityscapes dataset (top row),
             the Synscapes dataset (middle row) and our generated GANscapes
             dataset from the Synscapes semantic segmentation (last row).}
    \label{fig:dataset_examples}
\end{figure*}

To evaluate the proposed experiments real and synthetic datasets with
the same labeling specifications for ground truth labels are needed.
Cityscapes \cite{Cordts2016} was selected as the real world dataset and
the Synscapes dataset \cite{Wrenninge2018} with $N_\mathrm{syn}=25000$ examples
was chosen as the synthetic dataset.

\begin{figure}[thpb]
    \centering
    \parbox{2.9in}{
      \includegraphics[width=2.9in]{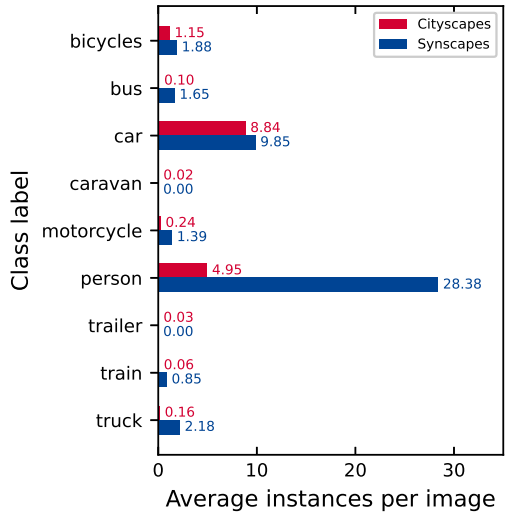}
    }
    \caption{Average instances per image for different classes in the
             Cityscapes and Synscapes/GANscapes dataset.}
    \label{fig:gt_instances}
\end{figure}

In addition to the labeling specifications,
many parameters in Synscapes like camera
pose and field of view are identical to the corresponding Cityscapes
parameters.
Some typical examples of Cityscapes and Synscapes images are shown in Fig.
\ref{fig:dataset_examples}.
The Synscapes dataset puts its emphasis on classes usually underrepresented
in autonomous driving datasets like persons.
Fig. \ref{fig:gt_instances} shows class imbalances of the two used datasets.
While in Cityscapes the car class is the dominant class, in Synscapes the
most frequent class is deliberately shifted towards the person class.

Since the Synscapes dataset does not distinguish between riders and rideable
objects (bicycles and motorcycles) a preprocessing step was performed on the
Cityscapes dataset where riders were merged with the closest rideable object
to the class bicycle or motorcycle.

The Cityscapes dataset contains 2975 labeled training images, 500 labeled 
validation images, and 1525 test images, for which the ground truth labels
are withheld for the official benchmark.
Therefore, the Hamburg sequence with 248 images was used as the new validation
set, which leaves 
$N_\mathrm{city}=2727$
examples in our real world training dataset.
The 500 original validation images were used as the new test set on which
all trained object detection networks were evaluated.

The GANscapes dataset was generated 
fom the semantic and instance segmentation images of the
Synscapes dataset with a pix2pixHD network 
\cite{Wang2018}.
For the experiments the original 
implementation\footnote{https://github.com/NVIDIA/pix2pixHD}
with weights from training on Cityscapes was used.
Since Synscapes does not differentiate between
the rider class and the bicycle or motorcycle class, 
the quality of GAN generated bicycles and motorcycles might be negatively
affected.
However, by looking at some generated bicycles and motorcycles (see e.g. last
row in Fig. \ref{fig:dataset_examples}), the quality is comparable to the
input from Cityscapes semantic and instance segmentation.
To further improve the quality of the generated images, the faces of semantic
segmentation covering the always visible front view of the recording car
are added to Synscapes semantic segmentation.

To analyze the detection performance of the trained networks with the
proposed power law, the networks have to be trained on a varying
(ideally logarithmically spaced) number of total training examples
$N_{r, i}$, where $r$ denotes the ratio of real world examples
and an index $i=1,\ldots, 10$.
These training subsets are constructed by randomly drawing
(without replacement) $r N_{r, i}$ examples from the $N_\mathrm{city}$
Cityscapes examples and $(1-r) N_{r, i}$ examples from the
$N_\mathrm{syn}$ synthetic examples.
For example, a training subset with
$N_{20\%, 1}=272$ examples consists of 54 examples randomly selected
from the $N_\mathrm{city}$ examples.
The remaining 218 examples are either selected from the $N_\mathrm{syn}$
Synscapes examples, when training on a mixed Cityscapes+Synscapes
dataset or selected from $N_\mathrm{syn}$ GANscapes examples when
training on a mixed Cityscapes+GANscapes dataset.

For the object detection network a YOLOv3 architecture\cite{Redmon2018}
was chosen, which was pretrained on the COCO dataset.
The YOLOv3 is trained with a batch size of 16, an initial learning rate of
$3 \cdot 10^{-4}$ and a cosine learning rate schedule without restarts
\cite{Loshchilov2017} is employed.
The training data is augmented via random color jitter, random crop, and
random mosaicking \cite{Bochkovskiy2020}.
The network is trained until no improvement on the validation set is
observed for 20 epochs.

For every number of training examples $N_{r,i}$
the training was repeated five times with different initializations
of the YOLOv3 detection head and different random selections of training
examples.

\section{RESULTS AND DISCUSSION}

\subsection{Pretraining + fine-tuning vs. mixed training strategy}

Results in a previous study \cite{Nowruzi2019} indicated that pretraining
a neural network on synthetic data and fine-tuning on real data is a
better training strategy than training on a mixed dataset of
synthetic and real data.
However, no confidence intervals are reported for their findings.
Therefore, a YOLOv3 network pretrained on COCO was on the one hand
pretrained on 25000 synthetic examples and subsequently fine-tuned
on 2727 real examples and on the other hand trained on a mixed dataset
consisting of all synthetic and real examples.
Both training strategies were repeated five times with different
random seeds.
The performance of the trained networks on the test set with 500 real examples
is evaluated in terms of mean average precision (mAP) and
average precision (AP) \cite{Everingham2010, Padilla2020}.
The metrics require an intersection over union (IoU) threshold
at which a predicted bounding box is considered a match with
a ground truth bounding box.
A threshold of $\mathrm{IoU} \geq 50\%$ is chosen throughout all experiments
(mAP$_{50}$, AP$_{50}$).
Tab. \ref{tab:mixed_vs_pretrain}
shows the resulting mAP$_{50}$ and AP$_{50}$ values
for the \emph{car} and \emph{person} class averaged over five training runs
with the standard deviation.
No significant difference between the pretraining + fine-tuning and the mixed
training strategy was observable.

\begin{table}[b]
    \begin{center}
        \caption{\textnormal{
        mAP$_{50}$ and AP$_{50}$ for the car and person class for both
        considered training strategies with a ratio of 10\% real data.}}
        \label{tab:mixed_vs_pretrain}
        \begin{tabular}{c c c}
        \hline\hline
                          & \bf{pretrain + fine-tune} & \bf{mixed} \\ \hline
        mAP$_{50}$        & 0.3706 $\pm$ 0.0082       & 0.3841 $\pm$ 0.0132 \\
        car AP$_{50}$     & 0.6356 $\pm$ 0.0065       & 0.6346 $\pm$ 0.0059 \\
        person AP$_{50}$  & 0.3518 $\pm$ 0.0128       & 0.3778 $\pm$ 0.0058 \\ \hline
        \end{tabular}
    \end{center}
\end{table}

\subsection{Reduction of real data amount by adding synthetic data}

\begin{figure*}[thb]
      \centering
      \parbox{5.5in}{
        \includegraphics[width=5.5in]{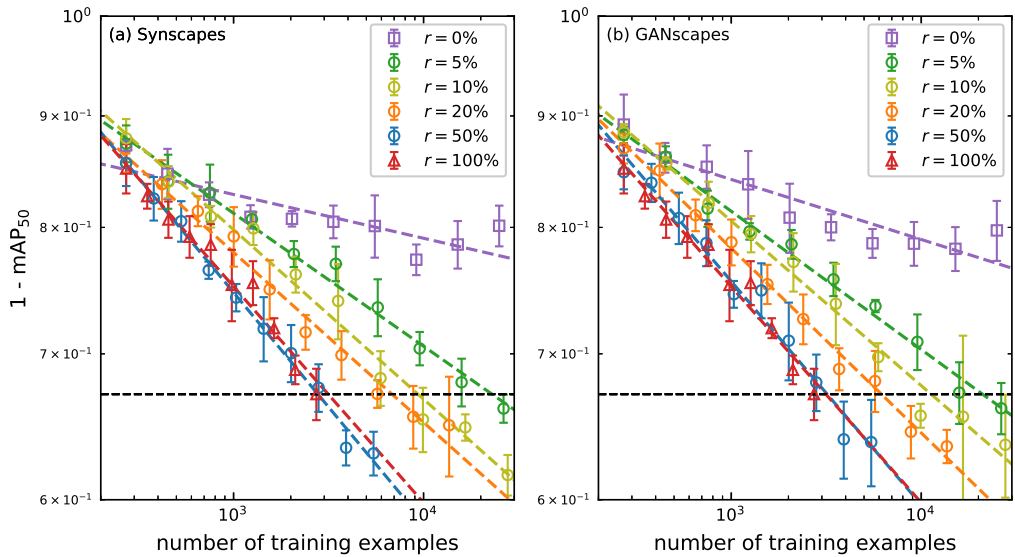}
      }
      \caption{Performance of neural networks in terms of 1-mAP$_{50}$ (lower
      is better) for various ratios of real data in the mixed dataset
      consisting of real data and Synscapes (left side) and
      GANscapes (right side). Values are presented as mean values and
      standard deviations over five training runs. Dashed colored lines
      are regression lines fitted with Eq. (\ref{eq:power_law}).
      The dashed horizontal black is an auxiliary line to help the reader
      compare detection performance with the mean performance of the networks
      trained on all 2727 real examples.}
      \label{fig:mAP_vs_training_examples}
\end{figure*}

As outlined in the introduction, neural networks trained only
on synthetic data generalize poorly to real world data.
Considering Fig. \ref{fig:mAP_vs_training_examples}, the previous
statement can be confirmed:
When the mAP$_{50}$ scores of networks trained on different ratios
of real and synthetic data are compared, networks trained on real data
only (red triangles, real data portion of 100\%)
outperform significantly their counterparts trained on synthetic
data only (purple squares, real data portion of 0\%).
These results are consistent for the Synscapes dataset
Fig. \ref{fig:mAP_vs_training_examples}a and the GANscapes dataset
Fig. \ref{fig:mAP_vs_training_examples}b.

In Fig. \ref{fig:mAP_vs_training_examples} we compare the model performance
for different ratios $r$ of real data dependent on the total number of
training examples.
In agreement with results by Hestness et al. the performance improvement
by increasing the number of training examples can be described by a power law
\cite{Hestness2017} of the form
\begin{equation}
    1 - \mathrm{mAP}_{50} = 10^\beta N_{r, i}^{\gamma},
\label{eq:power_law}
\end{equation}
where $\beta$ and $\gamma$
are the intercept and the slope of the linear fit in the log-log plot.
It can be seen in Fig. \ref{fig:mAP_vs_training_examples}, that the
slope parameter $\gamma$ decreases when ratio $r$ of real
data is increased, indicating better training results when more real data
is included in the training dataset.
The mAP$_{50}$ scores of networks trained on mixed datasets surpass
the best mAP$_{50}$ score for networks trained only on the real data
($r=100\%$) with the maximum number of training images
($N_{100\%, 10}=2727$).
However, by taking into account the standard deviations, the performance
increase is not significant in most cases.

The amount of training examples needed to reach the best performance
of the networks trained only on real data (intersection of colored dashed
lines with horizontal black dashed line in
Fig. \ref{fig:mAP_vs_training_examples}a and
Fig. \ref{fig:mAP_vs_training_examples}b)
can be easily obtained from Eq. (\ref{eq:power_law}).
The calculated numbers of total training examples and real training
examples are given in Tab. \ref{tab:data_reduction}.
While there is a substantial increase in the total number of required
training examples, the amount of real data examples decreases dramatically.
In the best case (10\% real images with Synscapes) only around 30\% of
the original real data were needed.
The results from Tab. \ref{tab:data_reduction} suggest that the best
ratio of real data in a mixed training dataset is between 5\% and 20\%.
The GANscapes dataset achieves comparable
results in terms of real data reduction.

\begin{figure*}[thb]
    \centering
    \parbox{5.5in}{
      \includegraphics[width=5.5in]{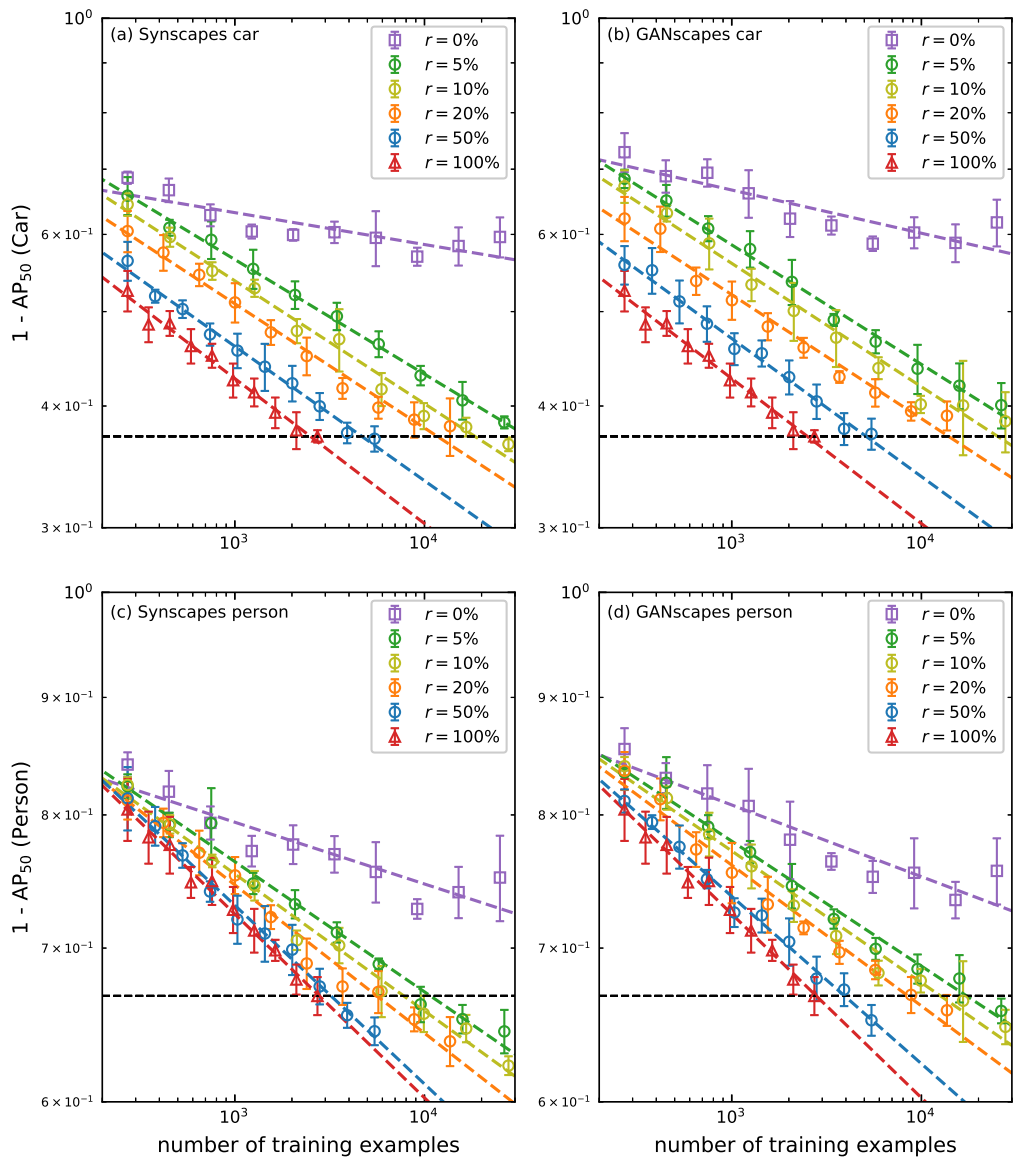}
    }
    \caption{Performance of neural networks in terms of 1-AP$_{50}$ (lower
    is better) for various ratios of real data in the mixed dataset
    consisting of real data and Synscapes (left side) and
    GANscapes (right side). Data is given for the car class (upper row)
    and the person class (lower row). Values are presented as mean values and
    standard deviations over five training runs. Dashed colored lines
    are regression lines fitted with Eq. (\ref{eq:power_law}).
    The dashed horizontal black is an auxiliary line to help the reader
    to compare the detection performances with the mean performance
    of the networks trained on all 2727 real examples.}
    \label{fig:AP_car_AP_person_vs_training_images}
\end{figure*}

\begin{table}[tb]
    \caption{\textnormal{
        Number of total images (synthetic + real images)
        needed to reach the performance of the model trained only on
        real data and the number of real images contained
        in the mixed dataset.
        Numbers are given for training with Synscapes and GANscapes
        images.}}
    \label{tab:data_reduction}
    \begin{center}
    \begin{tabular}{r r r r r}
    \hline\hline
                       & \multicolumn{2}{c}{\bf{Synscapes}} & \multicolumn{2}{c}{\bf{GANscapes}} \\
    \bf{\shortstack{Ratio of\\real data}} & \bf{\shortstack{Total\\images}} & \bf{\shortstack{Real\\images}} & \bf{\shortstack{Total\\images}} & \bf{\shortstack{Real\\images}}\\ \hline
      5\% & 18166 &  908 & 16804 &   840 \\
     10\% &  7783 &  778 &  9335 &   933 \\
     20\% &  5685 & 1137 &  5217 & 1043 \\
     50\% &  2427 & 1214 &  2783 & 1392 \\
    100\% &  2727 & 2727 &  2727 & 2727 \\ \hline   
    \end{tabular}
    \end{center}
\end{table}

\subsection{Class influence on data reduction}

The results shown for the mAP$_{50}$ score were obtained by averaging 
over all classes in the dataset, which masks the influence of specific
classes.
It is therefore interesting to analyze the detection performance for
different classes in terms of AP$_{50}$ score.
As an example the \emph{car} class (most frequent class in real dataset)
and the \emph{person} class (most frequent class in synthetic dataset)
were investigated in more detail.
From Fig. \ref{fig:AP_car_AP_person_vs_training_images}a and
Fig. \ref{fig:AP_car_AP_person_vs_training_images}b, it can be seen
that adding synthetic examples to the training dataset does not deteriorate
the detection performance of networks evaluated on the \emph{car}
class significantly.
Fig. \ref{fig:AP_car_AP_person_vs_training_images}c and
Fig. \ref{fig:AP_car_AP_person_vs_training_images}d show the 
same for the \emph{person} class.
The networks trained on mixed datasets exceed the detection
performance of the networks trained on real data for the \emph{person} class.
However, the performance increase is not significant except for the
training with $r=10\%$ with Synscapes data.

The real data saving results are presented in
Tab. \ref{tab:data_reduction_car}
for the \emph{car} class and in
Tab. \ref{tab:data_reduction_person}
for the \emph{person} class.
A large discrepancy between the \emph{car} and the 
\emph{person} class occurs when
the mixed datasets are compared in terms of real world data reduction.
For the \emph{car} class the lowest amount of real data examples 
required is still as much as 64\%
while for the \emph{person} class only 20\% of the real world dataset is
needed (5\% real with Synscapes).
This can be expected, because the real world dataset already contains a lot of
\emph{car} instances and hence the benefit by adding synthetic instances
is small.

Since the \emph{person} class is underrepresented in the real dataset,
the network benefits to a greater extent from the additional synthetic
instances.
Another interesting result is that the person assets, which were
generated in Synscapes with a lot of effort, can be exchanged with GAN
images without compromising too much reduction of real world data
examples in the training dataset.

\begin{table}[tb]
    \caption{\textnormal{
        Number of total images (synthetic + real images)
        needed to reach the AP$_{50}$ of the model trained only on
        real data for car class and the number of real images
        contained in the mixed dataset.
        Numbers are given for training with Synscapes and GANscapes
        images.}}
        \label{tab:data_reduction_car}
    \begin{center}
    \begin{tabular}{r r r r r}
    \hline\hline
    & \multicolumn{4}{c}{\bf{Car class}} \\
                       & \multicolumn{2}{c}{\bf{Synscapes}} & \multicolumn{2}{c}{\bf{GANscapes}} \\
    \bf{\shortstack{Ratio of\\real data}} & \bf{\shortstack{Total\\images}} & \bf{\shortstack{Real\\images}} & \bf{\shortstack{Total\\images}} & \bf{\shortstack{Real\\images}}\\ \hline
      5\% & 34887 & 1744 & 40059 & 2002 \\
     10\% & 18437 & 1844 & 25230 & 2523 \\
     20\% & 11683 & 2337 & 13901 & 2780 \\
     50\% &  4673 & 2337 &  5112 & 2556 \\
    100\% &  2727 & 2727 &  2727 & 2727 \\
    \hline   
    \end{tabular}
    \end{center}
\end{table}

\begin{table}[tb]
    \caption{\textnormal{
        Number of total images (synthetic + real images)
        needed to reach the AP$_{50}$ of the model trained only on
        real data for person class and the number of real images
        contained in the mixed dataset.
        Numbers are given for training with Synscapes and GANscapes
        images.}}
        \label{tab:data_reduction_person}
    \begin{center}
    \begin{tabular}{r r r r r}
    \hline\hline
    & \multicolumn{4}{c}{\bf{Person class}} \\
                       & \multicolumn{2}{c}{\bf{Synscapes}} & \multicolumn{2}{c}{\bf{GANscapes}} \\
    \bf{\shortstack{Ratio of\\real data}} & \bf{\shortstack{Total\\images}} & \bf{\shortstack{Real\\images}} & \bf{\shortstack{Total\\images}} & \bf{\shortstack{Real\\images}}\\ \hline
      5\% & 10752 &  538 & 17324 &  866 \\
     10\% &  7775 &  778 & 12620 & 1262 \\
     20\% &  5612 & 1122 &  8491 & 1698 \\
     50\% &  3214 & 1607 &  3935 & 1967 \\
    100\% &  2727 & 2727 &  2727 & 2727 \\
    \hline   
    \end{tabular}
    \end{center}
\end{table}

\section{CONCLUSIONS}

This study extends the research on the use of synthetic data for
neural network training.
In contrast to \cite{Nowruzi2019}, no significant difference of the detection
performance was found, when pretraining on synthetic data first
and fine-tuning on real data or when training on a mixed dataset of real and
synthetic examples.

We showed that synthetic data can be used to reduce the need for real
world data in a mixed training dataset.
The proposed method enabled us to estimate that the most reduction of
real examples can be achieved with a ratio of real
examples between 5\% and 20\% in mixed training datasets.
In addition, we showed that neural networks can particularly benefit
from synthetic data, when the synthetic data is enriched with classes that
are underrepresented the real world dataset.
Since synthetic data can usually be produced in a much more cost-efficient
way, mostly because ground truth labeling comes for free in synthetic
data, this is a promising approach for autonomous driving, where
real world labeled data is still rather expensive.

While synthetic datasets generated by GANs are not as good as
synthetic datasets produced by classical methods,
they are a promising alternative as the technology is still evolving.
Especially the need for modeling high fidelity 3D assets can be circumvented
by GANs. However, current GANs often need semantic (instance) segmentation
images for their training, which is even more expensive than
bounding box labeling.

Although the results are promising, our study only evaluated one
real dataset and two synthetic datasets for a single object detection
architecture.
Future work should therefore extend our proposed method for the evaluation
of reduction of real data in mixed training datasets to additional real and
synthetic datasets as well as additional object detection architectures.




\end{document}